\documentclass[sigconf]{acmart}
\AtBeginDocument{%
  \providecommand\BibTeX{{%
    \normalfont B\kern-0.5em{\scshape i\kern-0.25em b}\kern-0.8em\TeX}}}

\setcopyright{acmcopyright}
\copyrightyear{2023}
\acmYear{2023}
\setcopyright{acmlicensed}
\acmConference[CHI '23]{Proceedings of the 2023 CHI Conference on Human Factors in Computing Systems}{April 23--28, 2023}{Hamburg, Germany}
\acmBooktitle{Proceedings of the 2023 CHI Conference on Human Factors in Computing Systems (CHI '23), April 23--28, 2023, Hamburg, Germany}
\acmPrice{15.00}
\acmDOI{10.1145/3544548.3581373}
\acmISBN{978-1-4503-9421-5/23/04}

\usepackage{bbm}
\usepackage{dsfont}
\usepackage{xspace}
\usepackage{bm,amsmath}
\usepackage{booktabs}
\usepackage{multirow}

\begin{document}

\title{ESCAPE: Countering Systematic Errors from Machine's Blind Spots via Interactive Visual Analysis}

\author{Yongsu Ahn}
\authornote{This work was embarked during the author's internship at Bosch.}
\email{yongsu.ahn@pitt.edu}
\orcid{0000-0002-5797-5445}
\affiliation{%
  \institution{University of Pittsburgh}
  \city{Pittsburgh}
  \state{Pennsylvania}
  \country{United States}
}

\author{Yu-Ru Lin}
\email{yurulin@pitt.edu}
\affiliation{%
  \institution{University of Pittsburgh}
  \city{Pittsburgh}
  \state{Pennsylvania}
  \country{United States}
}

\author{Panpan Xu}
\email{xupanpan@amazon.com}
\orcid{0000-0002-8497-3015}
\affiliation{%
  \institution{Amazon AWS}
  \city{Santa Clara}
  \state{California}
  \country{United States}
}

\author{Zeng Dai}
\email{zengdai@meta.com}
\affiliation{%
  \institution{Meta}
  \city{Santa Clara}
  \state{California}
  \country{United States}
}

\renewcommand{\shortauthors}{Trovato and Tobin, et al.}

\newcommand{\ys}[1]{{\color{blue}{[YS: #1]}}}
\newcommand{\panpan}[1]{{\color{green}{[PX: #1]}}}
\newcommand{\rev}[1]{{\color{blue}{#1}}}
\newcommand{\yrl}[1]{{\color{red}{[YL: #1]}}}

\renewcommand{\vec}[1]{\mathbf{#1}}
\newcommand{\mat}[1]{\mathbf{#1}}

\newcommand{\name}{\texttt{ESCAPE}\xspace}
\newcommand{\namee}{\texttt{TribalGram}\xspace}

\newcommand{\diagnosisview}{\texttt{Misclassification Dignosis View}\xspace}
\newcommand{\performancechart}{\texttt{Performance Chart}\xspace}
\newcommand{\instancespace}{\texttt{Instance Space}\xspace}
\newcommand{\confidencescorefilter}{\texttt{Confidence Score Filter}\xspace}
\newcommand{\confusionmatrix}{\texttt{Confusion Matrix View}\xspace}
\newcommand{\confusionspace}{\texttt{Confusion Space}\xspace}
\newcommand{\contrastiveview}{\texttt{Contrastive Analysis View}\xspace}
\newcommand{\conceptlist}{\texttt{Concept List}\xspace}
\newcommand{\segmentview}{\texttt{Segment View}\xspace}

\newcommand{\conceptinspectionview}{\texttt{Concept Inspection View}\xspace}
\newcommand{\conceptassociationplot}{\texttt{Concept Association Plot}\xspace}
\newcommand{\conceptdetailview}{\texttt{Concept Detail View}\xspace}
\newcommand{\debiasview}{\texttt{Debias View}\xspace}
\newcommand{\debiasplot}{\texttt{Debias Plot}\xspace}

\begin{abstract}
Classification models learn to generalize the associations between data samples and their target classes. However, researchers have increasingly observed that machine learning practice easily leads to systematic errors in AI applications, a phenomenon referred to as ``AI blindspots.''  Such blindspots arise when a model is trained with training samples (e.g., cat/dog classification) where important patterns (e.g., black cats) are missing or periphery/undesirable patterns (e.g., dogs with grass background) are misleading towards a certain class. Even more sophisticated techniques cannot guarantee to capture, reason about, and prevent the spurious associations. In this work, we propose ESCAPE, a visual analytic system that promotes a human-in-the-loop workflow for countering systematic errors. By allowing human users to easily inspect spurious associations, the system facilitates users to spontaneously recognize concepts associated misclassifications and evaluate mitigation strategies that can reduce biased associations. We also propose two statistical approaches, relative concept association to better quantify the associations between a concept and instances, and debias method to mitigate spurious associations. We demonstrate the utility of our proposed ESCAPE system and statistical measures through extensive evaluation including quantitative experiments, usage scenarios, expert interviews, and controlled user experiments.
\end{abstract} 



\begin{teaserfigure}
  \includegraphics[width=\textwidth]{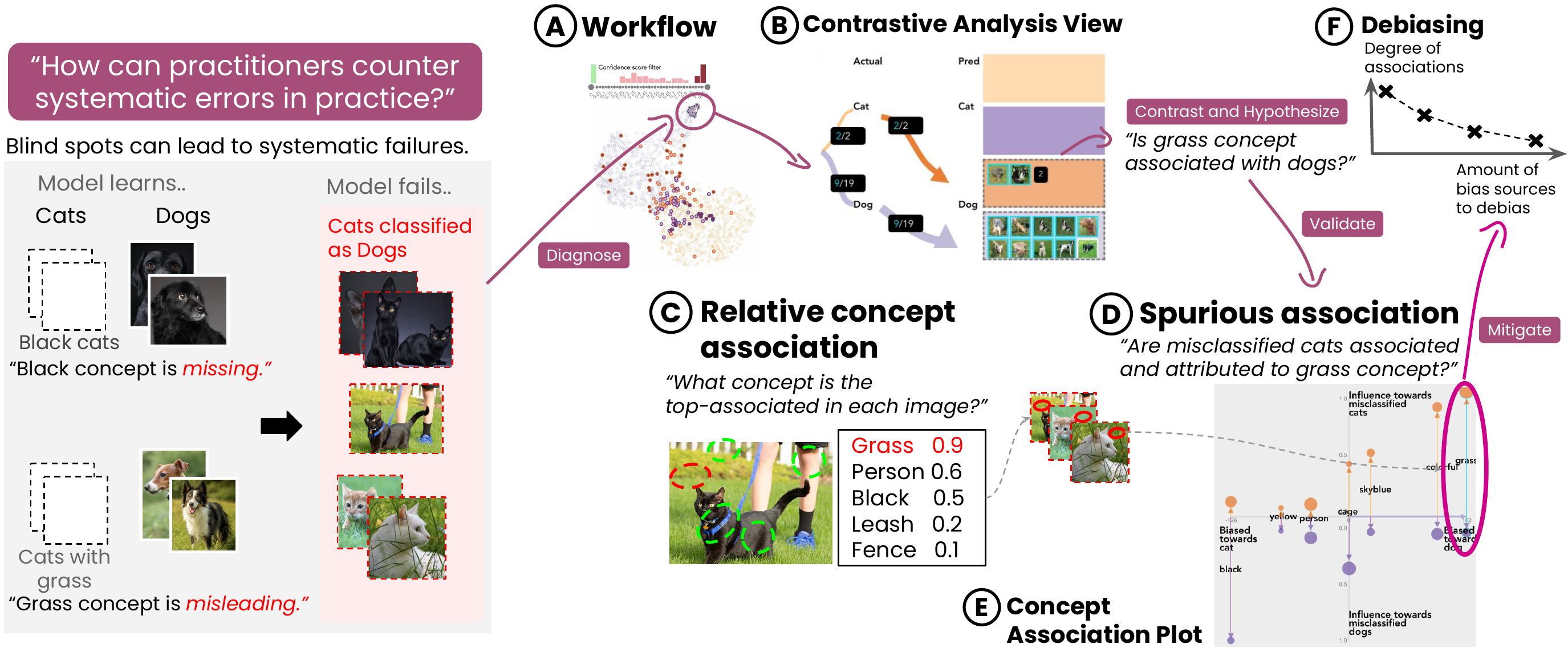}
  \vspace{-17.5pt}
  \caption{\label{fig:teaser} \name provides a prognostic workflow, methods, and visual components for countering systematic errors under a unified perspective of spurious associations. To tackle systematic errors induced from machine's blind spots (illustrated on the left side), our study promotes (A) an informed practice of a human-in-the-loop workflow where practitioners can diagnose, contrast, and hypothesize "missing" or "misleading" associations between an interpretable concept and a certain class (e.g., grass concept with dogs) from misclassifications via interactive modules. Users are further provided with (C, D, F) a set of proposed statistical methods and (E) a visual summary to validate and mitigate the associations if the target class is associated and attributed to a concept.}
\end{teaserfigure}

\maketitle

\section{Introduction}
\label{sec:introduction}
Classification models learn from data to generalize the associations between data samples and their target classes. However, researchers have increasingly observed that this process can easily lead to systematic errors. This phenomenon, also referred to as ``AI blindspots'' \cite{BeatMachineChallengingHumans, ContradictMachineHybridApproach}, arises when data samples do not well-represent the associations within the possible world of classes. These blind spots are difficult to prevent when training samples exhibit spontaneous patterns as undesired properties of target classes and thus create spurious associations.

\textbf{Motivating example.} To illustrate the problem, consider a simple scenario of an image classification model that predicts whether each image contains cats or dogs, as illustrated in Figure \ref{fig:teaser}. Suppose this model was trained with a dataset where (1) important/core patterns are \textit{missing} (e.g., whiteness or blackness) and (2) periphery/background/undesirable concepts (e.g., a dog with grass background) are \textit{misleading} due to biases towards a certain class. When deploying the model, test cases with these spurious associations will cause systematic errors (e.g., white dogs misclassified as cats, black cats, or cats with grass backgrounds misclassified as dogs). In high-stakes applications such as medical diagnosis or predictive policing, this leads to potential harm to individuals whose attributes are missing from the training set or associated with arbitrary patterns \cite{crawford2016there}.


Despite its consequences, such blind spots are difficult to detect, reason about, and prevent in practice. {While state-of-the-art techniques have proposed ways to detect blind spots \cite{coveragebasedutility, lakkaraju2017identifying} or underperforming subclasses and slices \cite{DominoDiscoveringSystematicErrorsCrossModal, SpotlightGeneralMethodDiscoveringSystematic}, they do not explain how misclassifications are \textit{attributed} to particular patterns, and thus provide little insight to overcome the blind spots. Recent XAI methods advancing concept-level interpretability \cite{AutomaticConceptbasedExplanations, InterpretabilityFeatureAttributionQuantitative, ConceptWhiteningInterpretable2020} also do not help make sense of an interpretation of those spurious associations (e.g., whether unexpected patterns were associated with certain classes and led to systematic failures). The standard machine learning pipeline lacks the ability to provide awareness and a deeper insight into the systematic errors or {\it biases} introduced by the training data and to recommend strategies to mitigate such biases. In practice, it is costly to explore and verify all possible patterns, many of which are noisy and not semantically meaningful using automated techniques. While recent visual analytics systems \cite{AutomaticConceptbasedExplanations, zhao2021humanintheloopextraction, kwon2022dash,  ConceptExplorerVisualAnalysisConcept} leverage the human ability to identify interpretable patterns, they do not support the inspection and interpretation of systematic errors or help mitigate spurious associations.

Given the profound impact of AI blindspots on a wide range of downstream applications, in this work, we aim to counter systematic errors by promoting an informed practice that reveals and reduces blind spots in the machine-learning pipeline. We propose a workflow to guide users in identifying, evaluating, and making sense of potential blind spots using analytical measures and visualization, as well as tools that help examine and reduce patterns associated with systematic errors. In the case of image classification, a {\it concept} is referred to as a higher-level feature representation that can be associated with a conceptual entity or a group of pixels in images that are semantically relevant to parts of an image (e.g., blackness, grass) \cite{AutomaticConceptbasedExplanations}. By allowing practitioners to see concept patterns associated with a target class in a biased manner, the workflow guides practitioners to generate hypotheses and evaluate strategies to overcome the machine's blind spots. Built on the recent discussion on different causes of systematic errors \cite{DominoExtractingComparingManipulating, SpotlightGeneralMethodDiscoveringSystematic}, our work further investigates two undesirable \textbf{spurious associations} arising from the model training process: the {\it missing} and {misleading} associations. As shown in Fig. \ref{fig:teaser}, missing associations refer to patterns that could have existed in a given class but are completely missed by the model; misleading associations refer to patterns that are disproportionately more likely to appear in an alternative class -- both contribute to the systematic errors the model produces. Furthermore, our workflow enables users to disentangle the model's systematic errors through an interactive analysis of two separate but interrelated explanation targets: 1) \textbf{association}: how is a hypothesized concept learned to be a pattern disproportionately associated with a particular class in the training process (i.e., are concepts biased towards a certain class?), and 2) \textbf{attribution}: how are misclassifications attributed to certain concepts? The spurious associations, once identified, can be evaluated to answer whether an association between a concept and a certain class can be mitigated by removing the source of bias. To facilitate such a workflow, we develop \name, a visual analytics system for countering \textbf{E}rrors from \textbf{S}purious \textbf{C}oncept \textbf{A}ssociations via interactive ins\textbf{PE}ction. Our system is designed to enable ML practitioners to make sense of spurious associations in countering systematic errors. The system incorporates a suite of visual components to help identify and hypothesize suspicious concept patterns as well as modules for validating the hypotheses and mitigating unwanted biases. To summarize, our contributions include:

\begin{itemize}
    \vspace{-0.5mm}
    \item \textbf{A prognostic workflow to guide the inspection of spurious associations.} We provide a range of tasks as a workflow to \textit{diagnose}, \textit{identify}, \textit{validate}, and \textit{mitigate} the associations between concepts and target classes. It further guides users to make sense of errors produced by spurious associations with recommendations to mitigate those errors.
    \item \textbf{\name, a visual analytic system for countering errors by integrating statistical and visual components to guide the inspection of spurious concept associations.} As a part of system design, we present two novel visualizations, \contrastiveview to help users capture the patterns that exclusively appear in certain classes by comparing confusion cases, and \conceptassociationplot which helps better make sense of two connected aspects of spurious associations.
    \item \textbf{A quantitative method to measure concept associations and debias spurious associations.} We propose a set of statistical metrics: 1) combined concept association: to quantify the degree of relative associations of concepts within each image, 2) between-class disparity: to measure the disparity of concept associations between concepts, and 3) debias method: to remove the spurious associations between a concept and a certain class.
    \item \textbf{Extensive evaluations.} We demonstrate the utility of our study through extensive evaluations including case studies, expert interviews, and controlled user experiments. Two usage scenarios showcase how diverse concepts can be detected in the image classification settings. We further evaluate how the design of our system allows users to effectively detect diverse concepts via controlled experiments.
    \vspace{-2mm}
\end{itemize}

\section{Related Work}

\subsection{Pattern discovery on systematic AI errors}

Systematic errors, sometimes coined as blind spots or unknown-unknowns \cite{BeatMachineChallengingHumans}, refer to model's failure over a group of instances that share similar semantics. There are various approaches for discovering such patterns, including algorithmic, human, or hybrid techniques.

A number of studies have shown that fully algorithmic techniques can help automatically discover unknown-unknowns \cite{lakkaraju2017identifying, coveragebasedutility}. Recently, several studies have also been proposed to advance the methods towards discovering automatic slices or subclasses that are semantically coherent \cite{DominoDiscoveringSystematicErrorsCrossModal, SpotlightGeneralMethodDiscoveringSystematic}, or to propose a framework for evaluating blindspot discovery methods in a unified manner \cite{EvaluatingSystemicErrorDetectionMethods}.

On the other hand, researchers have also explored how human intelligence can identify blind spots where automatic techniques alone do not work. Several studies \cite{BeatMachineChallengingHumans, ContradictMachineHybridApproach, HybridHumanAIWorkflowsUnknown, InvestigatingHumanMachineComplementarity} demonstrated that a well-designed crowdsourcing study can detect problematic instances. Hybrid workflows to leverage the abilities of both humans and machines \cite{HybridHumanAIWorkflowsUnknown, lakkaraju2017identifying, han2021iterative, chung2018unknownexamples} have also been explored throughout several studies in proposing collaborative human-AI workflow \cite{HybridHumanAIWorkflowsUnknown} or generating text descriptions \cite{han2021iterative} about spurious patterns.

While these studies demonstrate how human intelligence plays a significant role, tool support is still lacking to guide practitioners to inspect, identify, and mitigate systematic errors. In our study, we provide a workflow and systematic support for inspecting which systematic errors are attributed to interpretable concepts.

\subsection{Visual analytics for ML diagnostics}
Visual analytics tools in recent years have evolved to offer interactive ways for inspecting the machine learning process. In general, these tools aim to better visualize the predictive results in a model-agnostic manner or present the structure of the model in a model-specific way. Model-agnostic approaches propose to better visualize machine learning results regardless of model types. Many visualizations among them are largely designed on the grounds of confusion matrix as tree or flow diagram \cite{shen2020designing, VisualizingSurrogateDecisionTrees}, comparative visual design \cite{ManifoldModelAgnosticFrameworkInterpretation, ExplainExploreVisualExplorationMachine, olson2021contrastive, kaul2021improvingcounterfactuals, krause2017workflow}, radial \cite{VisualMethodsAnalyzingProbabilistic} or multi-axes based layout \cite{SquaresSupportingInteractivePerformance}. On the other hand, model-specific inspections also gained attention to support the inspection of a deep neural network inside its layers, neurons, or activations \cite{liu2017analyzingtraining, ShapeShopUnderstandingDeepLearning, TopoActVisuallyExploringShape, DeepVIDDeepVisualInterpretation}.

Visual analytic tools can also help inspect and explain the potential cause of systematic failures such as a shifted or skewed distribution of the training examples termed as out-of-distribution \cite{OoDAnalyzerInteractiveAnalysisOutofDistribution}, covariate or concept shift \cite{DiagnosingConceptDriftVisual} or machine biases \cite{FairVisVisualAnalyticsDiscovering, FairSightVisualAnalyticsFairness, WhatIfToolInteractiveProbing}. The OoD analyzer \cite{OoDAnalyzerInteractiveAnalysisOutofDistribution} presented a grid-based layout to visualize the distributional differences in training and test sets. The problem of concept drift was tackled and presented as visualizations in a 2D heatmap visualization \cite{DiagnosingConceptDriftVisual} or distribution-based scatterplot \cite{ConceptExplorerVisualAnalysisConcept}. Other interactive tools such as Deblinder \cite{DiscoveringValidatingAIErrorsCrowdsourced}, SEAL \cite{SEALInteractiveToolSystematicError}, or Error Analysis \cite{erroranalysis} have recently been proposed to mitigate systematic errors with subclass labeling or user-generated report. Compared to previous work, our study aims to promote a human-in-the-loop workflow consisting of tasks to identify biased patterns and their association/attribution aspects with the perspective of spurious associations.


\subsection{Understanding model with concept interpretability}

The XAI methods to explain the behavior of black box models \cite{InterpretabilityFeatureAttributionQuantitative, AutomaticConceptbasedExplanations, BayesianCaseModelGenerative, ConceptWhiteningInterpretable2020} have been recently expanded to a concept-level sensitivity. The method called TCAV (Testing Concept Activation Vector) \cite{InterpretabilityFeatureAttributionQuantitative} provides a post-hoc method to explain the global influence of a concept in a pre-trained model. ACE (Automatic Concept Extraction) \cite{AutomaticConceptbasedExplanations} was proposed to identify and filter interpretable concepts from the meaningful clusters of segments on the basis of TCAV. In \cite{ConceptWhiteningInterpretable2020}, Concept Whitening (CW) purposefully alters batch normalization layers to a concept whitening layer to learn an interpretable latent space. Especially, the whitening step in this method points out that the concept space needs to be preprocessed to better align concept vectors.

These concept-level interpretability methods, however, require the human ability to observe and extract semantically meaningful concepts \cite{AutomaticConceptbasedExplanations}. There are various ways to identify and extract concepts in collaboration with humans and systems \cite{AutomaticConceptbasedExplanations, NeuroCartographyScalableAutomaticVisualSummarization, zhao2021humanintheloopextraction, DASHVisualAnalyticsDebiasingImage,  ConceptExplorerVisualAnalysisConcept, ProtoSteerSteeringDeepSequence, AnchorVizFacilitatingSemanticData, ConceptVectorTextVisualAnalytics, VisualConceptProgrammingVisualAnalytics}. ConceptExtract \cite{zhao2021humanintheloopextraction} aimed to support concept extraction and classification in a human-in-the-loop workflow and visual tools. In DASH \cite{kwon2022dash}, problematic biases from irrelevant concepts can be identified through observations from users, which were proposed to be mitigated through random image generation using GAN techniques. ConceptExplainer \cite{ConceptExplorerVisualAnalysisConcept} was designed to explore the concept associations focusing on validating conceptual overlapping between classes, especially serving as a concept exploration tool for non-expert users. In \cite{VisualConceptProgrammingVisualAnalytics}, a self-supervised technique was proposed to automatically extract visual vocabulary to allow experts to refine the labeled data and understand the concepts.

Unlike existing work, our study proposes an interactive workflow of exploring concepts for the purpose of inspecting systematic errors and spurious concept associations behind them. Similar to \cite{WhatDidMyAILearn}, our human-in-the-loop workflow aims to promote the sensemaking of practitioners specifically in the problem of systematic errors where they can iteratively work on subsetting, contrasting patterns in instances, and hypothesizing spurious associations.


\section{Challenges}
\label{sec:challenges}
In this section, we summarize several challenges in inspecting and making sense of systematic errors. This is based on the review of literature \cite{BeatMachineChallengingHumans, HybridHumanAIWorkflowsUnknown, ContradictMachineHybridApproach, SpotlightGeneralMethodDiscoveringSystematic, DominoExtractingComparingManipulating, WhyShouldTrustYou, InterpretabilityFeatureAttributionQuantitative, DASHVisualAnalyticsDebiasingImage} that tackle pattern discovery in systematic errors and post-hoc explainability methods by examining their research problem, limitations, and future work. For each of the four challenges (C1-C4), we also identify the need for tool support due to the lack of capability in existing methods otherwise it is challenging for practitioners to achieve the task: \\\vspace{-5pt}

\hangindent=2em \textbf{C1. Systematic errors are not easily diagnosable.} Systematic errors are critical but hardly recognizable among thousands of instances in high-dimensional representational spaces \cite{BeatMachineChallengingHumans, HybridHumanAIWorkflowsUnknown}. It is important to support practitioners (1) to narrow down the scope of classes that are prone to systematic errors and (2) to inspect errors with respect to the model's confidence, to allow users to diagnose the potential blind spots. \\\vspace{-5pt}

\hangindent=2em \textbf{C2. Practitioners cannot easily discover how interpretable concepts are associated with errors or test hypotheses based on their knowledge or observations.} Humans are known to classify images with semantic concepts \cite{HybridHumanAIWorkflowsUnknown, BeatMachineChallengingHumans, ContradictMachineHybridApproach}. However, existing tools rarely leverage this ability to discover and hypothesize concept associations against misclassifications \cite{SpotlightGeneralMethodDiscoveringSystematic, DominoExtractingComparingManipulating}. To hypothesize and validate the association, practitioners should be able to easily capture patterns and concepts that exclusively appear over misclassifications in a certain class. The hypothesized concepts need to be readily defined to be tested. \\\vspace{-5pt}

\hangindent=2em \textbf{C3. Spurious associations are not immediately recognizable and understandable, with few clues for intervention.} Once an association is hypothesized, users need to (1) make sense of how concepts are associated with certain classes then cause misclassifications in another class and (2) quantitatively verify the degree of associations and attributions. While these aspects of systematic errors were separately studied \cite{DominoExtractingComparingManipulating, WhyShouldTrustYou, InterpretabilityFeatureAttributionQuantitative} or have not been explored in existing work, practitioners should be able to understand what causes systematic errors and precisely verify spurious associations in an explainable and measurable manner. \\\vspace{-5pt}
    
\hangindent=2em \textbf{C4. The sources of biases in spurious associations are not traceable and and thus the biases are difficult to mitigate.} The spurious associations, once detected, should be mitigated with strategies to correct biased concept patterns and be properly evaluated as to whether the association is mitigated as expected. While most of studies \cite{DASHVisualAnalyticsDebiasingImage, DiscoveringValidatingAIErrorsCrowdsourced, BeatMachineChallengingHumans, HybridHumanAIWorkflowsUnknown, ContradictMachineHybridApproach, DominoExtractingComparingManipulatinga} do not provide a way to efficiently trace back to the source of biases in training instances or in-the-wild datasets, practitioners are still required to find a way to precisely determine and evaluate mitigation strategies in practice.
\begin{figure*}[!ht]
    \centering
    \includegraphics[width=0.95\linewidth]{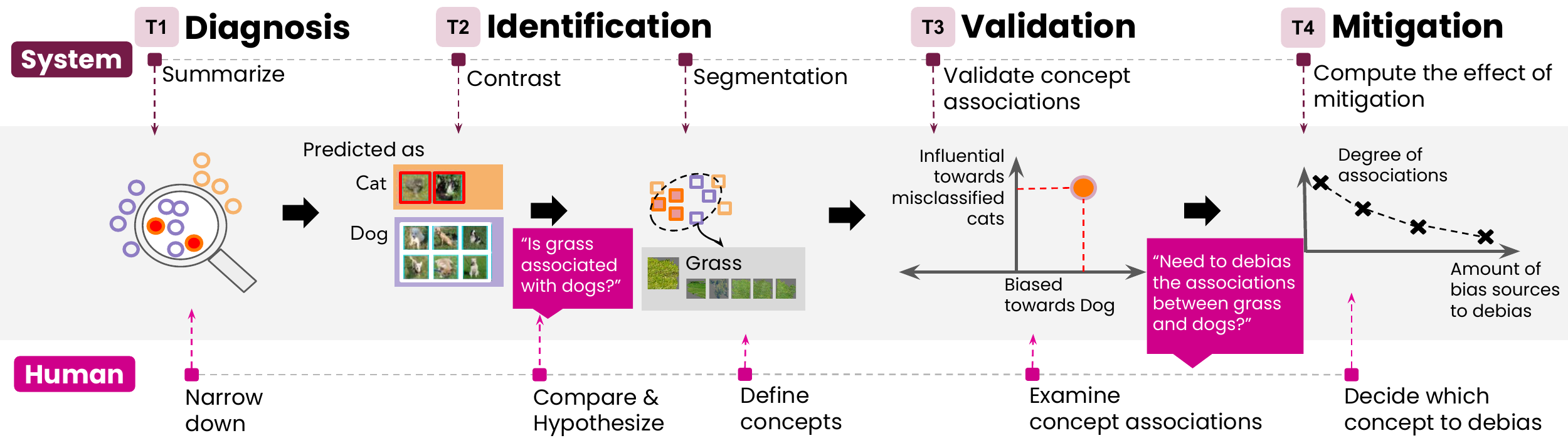}
    \caption{\label{fig:workflow}
    \textbf{The proposed workflow to guide the inspection of systematic errors and spurious concept associations.} A user starts with (T1) inspecting the degree of misclassifications to narrow down the scope of analysis to the potential regions with systematic errors. (T2) By observing interpretable concepts that exclusively appear in a certain class, users are provided to easily identify and hypothesize spurious associations between concepts and target classes. (T3) User-generated hypotheses can be validated via two aspects of systematic errors, associations, and attribution. (T4) Undesirable associations can be mitigated with a debias strategy in the system and evaluated how it helped mitigate the concept-class associations.
    }
\end{figure*}

\section{Workflow}

To address the challenges, we propose a workflow supported by our system to guide the inspection of systematic errors and spurious concept associations. Our workflow (Fig. \ref{fig:workflow}) consists of a range of four guided tasks (T1-T4) to diagnose, identify, validate, and mitigate the undesirable associations between concepts and target classes. We also describe how our system facilitates these tasks to address the aforementioned challenges in existing tools. \\

\hangindent=2em \textbf{T1. Diagnosis phase. Inspect the degree of misclassifications at the instance-level and class-level.} In Diagnosis phase (Fig. \ref{fig:workflow}-T1), users are guided to inspect the degree of misclassifications and model confidence and narrow down the scope of inspection to a subset of classes and instances that are vulnerable to systematic failures. To highlight systematic errors (C1), \name summarizes the degree of misclassifications and model confidence at the instance-level and class-level in the diagnostic modules highlighting misclassifications and unknown-unknowns. \\\vspace{-5pt}
    
\hangindent=2em \textbf{T2. Identification phase. Identify and hypothesize a biased association between patterns and target classes.} In Identification phase (Fig. \ref{fig:workflow}-T2), users are allowed to explore the patterns to hypothesize the associations between patterns and target classes. To support the task of hypothesizing suspicious patterns (C2), the system provides visual components that facilitate the comparison of patterns in different confusion cases. This helps users observe any patterns or concepts that are exclusively associated with a certain class. To test their hypotheses, users are provided with system modules to easily define a group of semantically coherent segments as concepts using the segmentation module in the system. \\\vspace{-5pt}
    
\hangindent=2em \textbf{T3. Validation phase. Verify the hypotheses over spurious concept association.} 
    In Validation phase (Fig. \ref{fig:workflow}-T3), users test their hypotheses by checking how concepts are associated with a certain class and influential towards misclassifications (C3). The system supports two-sided validations, associations, and attribution (details in Section \ref{sec:introduction}) quantified with statistical metrics described in Section \ref{sec:method}. These two aspects of associations are visualized for users to intuitively capture the concept associations and inspect the top-associated instances. \\\vspace{-5pt}
    
\hangindent=2em \textbf{T4. Mitigation phase. Remove the source of biases and mitigate the spurious associations.} In Mitigation phase (Fig. \ref{fig:workflow}-T4), users determine whether spurious associations between a concept and classes should be removed. When a strategy to mitigate the biases is performed, they need to evaluate whether spurious associations were removed (C4). In the system, we provide a statistical method to debias the training samples that are the most associated with a concept by neutralizing the representation of target instances via projection. In the system, this mitigation process is supported with interactive modules to visualize the effect of mitigation and help users determine the debiasing process.
\section{Measuring and mitigating spurious concept association via concept space}
\label{sec:method}

In this section, we present statistical approaches (1) to measure two aspects of systematic error: the associations between a concept and instances (especially misclassifications) (Section \ref{sec:method-concept-association}) and the attribution of predictions to certain concepts (Section \ref{sec:method-concept-influence}) for Validation phase (\textbf{T3}), and (2) to mitigate spurious associations (Section \ref{sec:method-debiasing}) in Mitigation phase (\textbf{T4}).

\subsection{Measuring concept associations}
\label{sec:method-concept-association}
First, we present a method of measuring the association between a concept and instances. We start by defining two different levels of associations, (1) Instance-level association: It quantifies how a concept is closely aligned with an individual instance at a fine-grained level. (2) Class-level association: By aggregating instance-level associations into classes, we compute how a concept is associated with a group of instances in each target class. This measure can be extended to quantify the between-class disparity over concept associations.

To compute these measures, we take the well-known perspective of concept space where the association between two subspaces, concepts, and instances,  can be measured. Suppose that $N$ images are given in a machine learning task using a deep classifier $f$ consisting of a group of layers. Given a specific layer $l$ of interest, a subset of images $X$ and segments $S$ are vectors $\{\vec{v}_x\}_{x \in X}$ and $\{\vec{v}_s\}_{s \in S}$ in the space of activations of layer $l$. An interpretable concept $c$ is represented as a group of segment images $S_c$ that share the semantics of the target concept. We define a concept vector $\vec{v_{c}}$ as a mean vector of a group of segment vectors (e.g., two segment sets colored red and yellow in Fig. \ref{fig:method-subspace-alignment}a).

\subsubsection{Instance-level concept association} First, we define the instance-level association between a concept $c$ and images $\vec{X}$ in the space of layer $l$. Given an image vector $\vec{v}_x$ and a concept vector $\vec{v}_c$, the association can be computed via cosine similarity:

\begin{equation}
    \label{equation:concept-association}
    CA_{(x, c)} = \frac{\vec{v}_x \times \vec{v}_c}{\|\vec{v}_x\| \|\vec{v}_c\| }
\end{equation}

\begin{figure}[!ht]
    \centering
    \includegraphics[width=0.95\columnwidth]{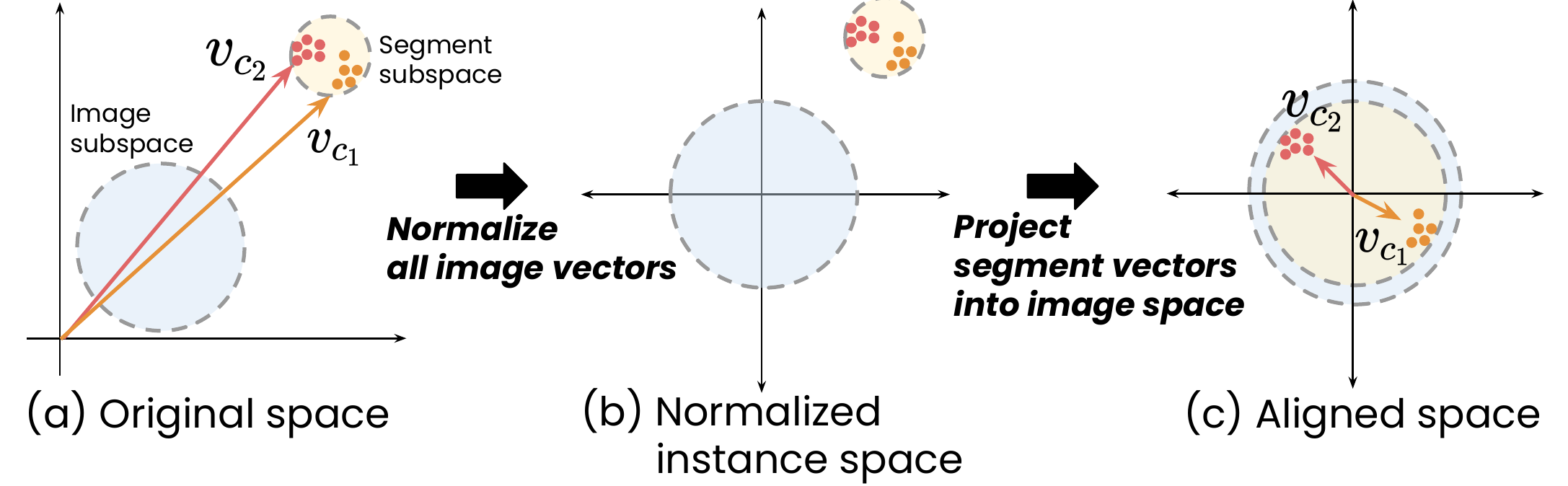}
    \vspace{-12.5pt}
    \caption{\label{fig:method-subspace-alignment}
    \textbf{The preprocessing steps for aligning instances and segments.} In the original space (a), two different concepts are not well-distinguishable. For better alignment, (b) the image subspace is normalized as shown, and (c) the segment subspace is then projected into the normalized image space. After preprocessing, two concept vectors are better discriminated in their directions.
    }
    \vspace{-2.5pt}
\end{figure}

\textbf{Subspace alignment.} These concept associations are measured via pairwise cosine similarity as angles between images $\{\vec{v_x}\}_{x \in X}$ and a concept $\vec{c}$ that seemingly work well; however, they are not distinguishable (Fig. \ref{fig:method-subspace-alignment}a), as pointed out in \cite{ConceptWhiteningInterpretable2020}. In our case, two subspaces of instances and segments are misaligned and placed far away from each other, so that two distinct concepts in Fig. \ref{fig:method-subspace-alignment}a result in indistinguishable cosine similarities, thus misleading associations between instances and concepts.

To address the misalignment problem, we perform the preprocessing to make two spaces better align with each other, consisting of two steps as illustrated in Fig. \ref{fig:method-subspace-alignment}: First, we normalize all instances to project them into the standardized space. Second, to align the segment space with the instance space, we project segment vectors into the normalized instance space using mean and standard deviation of instance vectors. As a result of the preprocessing steps, two concept vectors are well-distinguishable in their directions as much as they are semantically different (Fig. \ref{fig:method-subspace-alignment}c).

\textbf{Exclusive concept association.} After the preprocessing for subspace alignment, we compute exclusive concept association. This measure identifies which concepts within each image are relatively more influential than others as the top-associated concept.

\begin{figure}[!ht]
    \centering
    \includegraphics[width=\columnwidth]{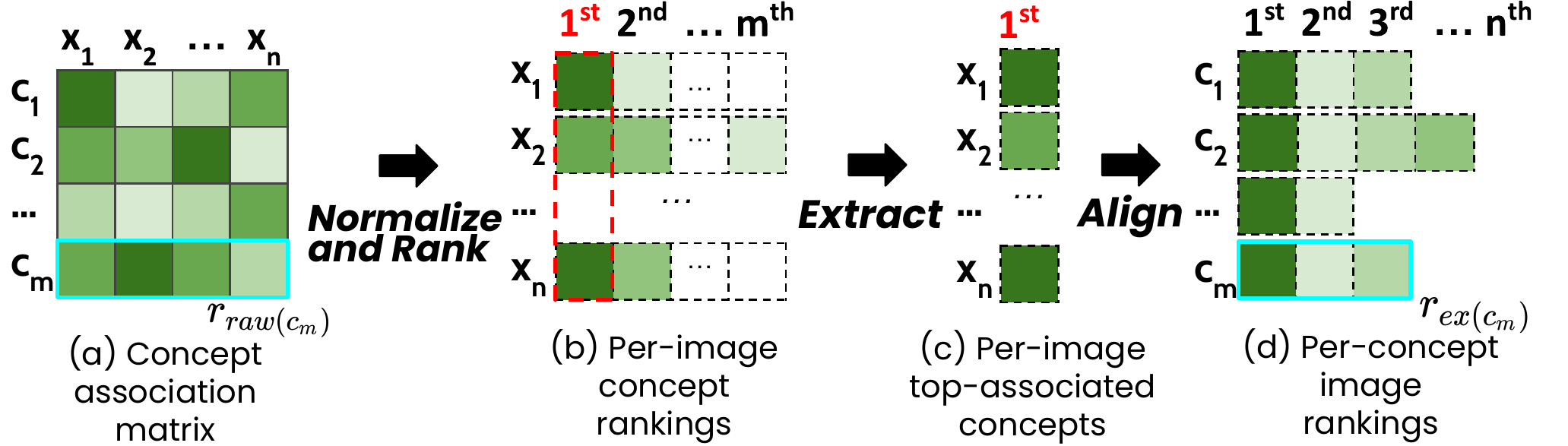}
    \caption{\label{fig:method-ex}
    \textbf{Three detailed steps for measuring exclusive concept association.}
    }
\end{figure}

Specifically, we take the following steps to compute exclusive association scores: (1) Measuring absolute concept associations: for a group of instances $X$ and concepts $C$, we calculate the pairwise cosine similarity (Equation \ref{equation:concept-association}) between $\{\vec{v_x}\}_{x \in X}$ and $\{\vec{v_c}\}_{c \in C}$ to measure how a concept and instance is associated in an absolute manner (Fig. \ref{fig:method-ex}a). From this matrix, by ranking instances in order of absolute scores for each row, we can obtain absolute concept associations as a set of per-concept image rankings $R_{raw(C)}=\{r_{raw(c)}\}_{c \in C}$. (2) Normalizing concept associations per image: for each image, association scores are standardized to convert them to relative association scores (i.e., column-wise standardization of the matrix in Fig. \ref{fig:method-ex}a). From the normalized matrix, each row as relative concept associations for each image is converted to a ranking, resulting in a set of per-image concept rankings (Fig. \ref{fig:method-ex}b). (3) Extracting the most associated concepts: From the per-image rankings, we extract the most-associated concept in each image, resulting in a obtain a list of top-associated concepts for each instance $\{(x, c_{top(x)})\}$. (4) Ranking the most-associated images per concept: These pairs are further grouped by each concept whose images are in the order of associations, resulting in a set of per-concept image rankings (Fig. \ref{fig:method-ex}d). These rankings $R_{ex(C)}=\{r_{ex(c)}\}_{c \in C}$ indicate how a concept is exclusively associated with a set of instances. 

\textbf{Combined association score.} Given two types of ranked associations, $R_{raw(C)}$ and $R_{ex(C)}$, we aim to combine them to find out instances that are the most associated with a concept absolutely and exclusively than other concepts. We use FREX score \cite{bischof2012summarizingfrex}, which was originally introduced in topic modeling, to combine raw and exclusive associations $R_{raw(C)}$ and $R_{ex(C)}$ across topic words:

\begin{equation}
    CA_{comb} = (\frac{w}{ECDF_{R_{ex}}} + \frac{1-w} {ECDF_{R_{raw}}})^{-1}
    \label{equation:combined-association-score}
\end{equation}

where $w$ is the weight between 0 and 1 given to exclusivity and ECDF is the empirical CDF function. We set $w=0.2$ to combine two rankings in our setting. Based on FREX score, we derive the combined ranking $R_{comb}$ to identify the list of concept-associated instances for the subsequent tasks.

\subsubsection{Class-level concept association and Between-class disparity}
\label{sec:method-between-class-disparity}

Based on the underlying measure of combined concept associations in Equation \ref{equation:combined-association-score}, we compute the between-class disparity to measure how a concept is more biased towards a certain class to compare the class-wise magnitude of concept associations. In the system, we use this metric over training examples to identify concept-wise biases towards classes in the training phase. It allows us to trace back to which concepts were learned to be patterns of one class relative to another. The between-class disparity is computed via taking the difference between the sum of combined concept associations based on the following equation:

\begin{equation}
    \label{equation:between-disparity}
    Disparity_{(c,X)} = \sum_{x \in X_{pos}}{CA_{comb(c,x)}} - \sum_{x \in X_{neg}}{CA_{comb(c,x)}}
\end{equation}

where $X_{neg}$ and $X_{pos}$ are two sets of instances belonging to either positive and negative class. When the difference is greater than zero, it indicates the concept is biased towards the positive class, otherwise negative class.

\subsection{Quantifying concept influences}
\label{sec:method-concept-influence}
Next, we quantify the influence of concepts on certain predictions, especially to measure how false predictions are attributed to a specific concept. We employ the approach of TCAV \cite{InterpretabilityFeatureAttributionQuantitative}, which uses directional derivatives to quantify how predictions over a group of instance vectors $\{\vec{v_x}\}_{x \in X}$ towards a certain class are sensitive to the direction of a concept vector $\vec{v_c}$. Compared to TCAV leveraging a concept activation vector (CAV) as an orthogonal direction of a hyperplane of trained linear classifier, we identify a concept vector $\vec{v_c}$ from a centroid of segment vectors $\vec{v_S}/\|S\|$, which is cost-effective in computations without having to train linear classifiers multiple times.

\subsection{Debiasing spurious concept associations}
\label{sec:method-debiasing}
In the workflow, the Mitigation phase aims to reduce the degree of spurious associations between a certain concept and target class. This requires a strategy to correct the spurious concept-class associations. While there are various approaches, such as active learning (i.e., adding training examples with desirable patterns), our goal is to prevent any undesirable concepts from being associated with a certain class. In this context, we propose a method to debias the proportion of concepts from instances that are highly associated with those concepts. We take the approach of debiasing from the line of work in fair representation \cite{sutton2018biased, bolukbasi2016man, zhao2018learning} (e.g., removing the gender bias from gender-specific words). For internal representations of instances containing undesirable attributes, it projects the target instances into the orthogonal direction of the unwanted attribute such as job or gender to neutralize them. This approach is applicable to our context to remove the property of a concept (as a vector $\vec{v_c}$) from each concept-associated instance (as a vector $\vec{v_i}, i \in I$). We apply this step to a group of training samples in $r_{comb(c)}$ that are identified as the most associated with a concept $c$:

\begin{equation}
    Debias_{(i,c)}: {\vec{v_i}} - \frac{{\vec{v_i}}*{\vec{v_c}}}{\|{\vec{v_c}} \times {\vec{v_c}}\| } \times {\vec{v_c}}
    \label{equation:debias}
\end{equation}

After this step, we compute the remaining bias ratio, the degree of the between-class disparity in Equation \ref{equation:between-disparity} being mitigated before and after debiasing:

\begin{equation}
    \label{equation:remaining-bias-ratio}
    Remaining \ Bias \ Ratio_{(c, X)} = 1 - \frac{(Disparity_{AF(c, X)} - Disparity_{BF(c, X)})}{Disparity_{BF(c, X)}}
\end{equation}

\section{System Components}
\label{sec:system}
\begin{figure*}[!ht]
    \centering
    \includegraphics[width=0.95\linewidth]{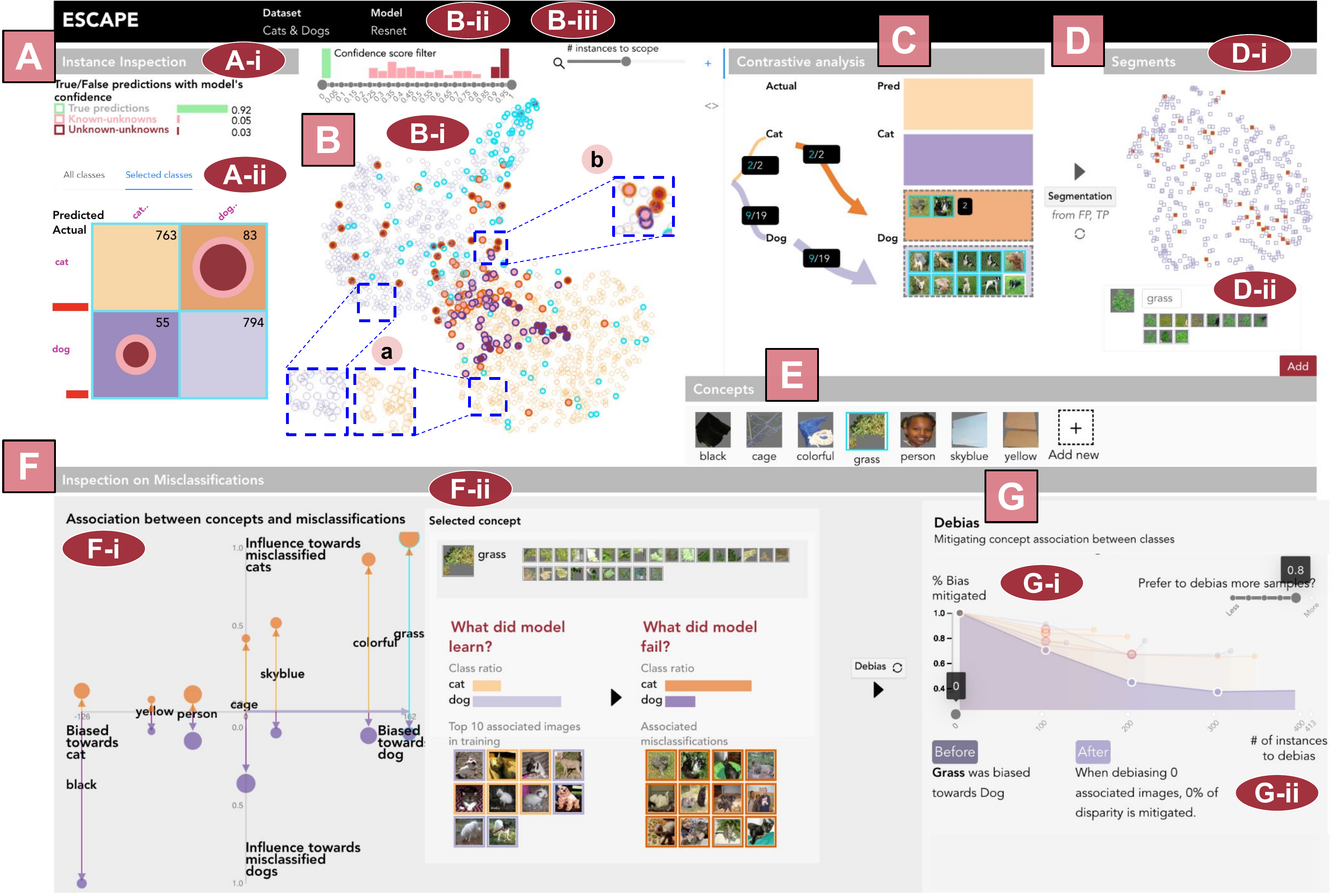}
    \caption{\label{fig:system}
    \textbf{The overview of \name system.} Once the system is initialized, (A) \diagnosisview provides the overview in the degree of misclassifications and model confidence. By selecting a pair of classes, (B) all instances are visually highlighted and can be filtered and observed for their patterns. (C) A group of instances in a focal area are visually contrasted in their true and predicted classes, which can be (D) segmented and (E) defined with a group of semantically coherent segments as a concept. (F) All defined concepts are visually represented in terms of how each of them is biased towards a certain class in the training set and influential towards misclassifications. (G) Once an undesirable concept-class association is identified, users can determine whether to attempt a debiasing strategy to remove the proportion of concepts from the training set and evaluate whether it has mitigated the associations.
    }
    \vspace{-5pt}
\end{figure*}

\name is designed to support the aforementioned workflow of inspecting spurious concept associations by incorporating a suite of visual interfaces and proposed statistical methods, which consists of several components, as illustrated in Fig. \ref{fig:system}A-G. Throughout the system, multiple color schemes are applied: (1) two true negative and positive classes are represented as orange and purple. These two colors are further diverged into lighter or darker orange/purple to differentiate four confusion cases (Fig. \ref{fig:system}A-ii). (2) Two types of misclassifications are presented as red colors, either pink or dark red to indicate low or high confidence that is referred to as known-unknowns and unknown-unknowns respectively. 

\subsection{Misclassification Diagnosis View}
\label{sec:misclassification-diagnosis-view}
\diagnosisview (Fig. \ref{fig:system}A) provides the diagnostic modules (\textbf{T1}) for summarizing the degree of misclassification at instance-level in \performancechart (Fig. \ref{fig:system}A-i) and class-level in \confusionmatrix (Fig. \ref{fig:system}A-ii). 

\confusionmatrix (Fig. \ref{fig:system}A-ii) is a confusion matrix visualization that is designed to highlight misclassifications and unknown-unknowns. In this view, false confusion cases are highlighted with two overlapping circles, consisting of a pink outer circle indicating (1) the number of misclassifications and dark red inner circle indicating (2) unknown-unknowns per class. The number of unknown-unknowns is presented per class as red bars along with row labels in the matrix. In the system, there are two \confusionmatrix s presented in a tab interface, representing all classes or a pair of selected binary cases. In \name, the rest of components are provided to support the inspection of a selected binary case. This design choice serves two objectives: (1) to highlight systematic errors that rarely appear without using visual spaces to represent all cases, and (2) to allow users to perform a careful inspection on spurious associations between two classes without the cognitive overload.

\subsection{Instance Space} \label{sec:instance-space}
\instancespace (Fig. \ref{fig:system}B-i) allows users to inspect the patterns of instances in a pair of classes (\textbf{T2}) selected from \confusionmatrix. In a two-dimensional scatter plot, all instances in the test set are represented as a circle, whose high-dimensional vectorized representations are projected into a 2D space using UMAP \cite{mcinnes2018umap}. In this view, all instances are represented with respect to their true and predicted cases either as hollow circles for true predictions (Fig.\ref{fig:system}B-i-a) or as filled circles for false predictions with its lighter/darker red color indicating low or high confidence in misclassifications (Fig.\ref{fig:system}B-i-b). These instance circles are stroked with either orange or purple based on actual classes (Fig.\ref{fig:system}B-i-a). In combination with two visual clues, misclassifications in two classes are visually highlighted as filled and bigger circles over \instancespace. When users hover over each instance circle, a tooltip shows the actual image of the instance, with its k nearest neighbors with k adjustable with a slider interface (Fig. \ref{fig:system}B-ii). 

These instances can be filtered based on confusion cases in \confusionmatrix by clicking on desirable cases and model confidence with \confidencescorefilter (Fig. \ref{fig:system}B-ii), which summarizes the distribution of unknown-unknown scores. We use brier score \cite{brier1950verification} to compute the discrepancy between a predictive distribution and true distribution in each instance, ranging from 0 (true prediction) to 1 (largest discrepancy in error). A group of instances can be selected by clicking on a region to appear in \contrastiveview.

\subsection{Contrastive Analysis View}
\label{sec:contrastive-analysis-view}

\contrastiveview (Fig. \ref{fig:system}C) allows users to compare the patterns of a group of instances within different confusion cases. Once selected in \instancespace, a group of instances are displayed in \contrastiveview. In a flow-based layout, four confusion cases are presented as an arrow connecting from actual to predicted class, with its thickness indicating the size of instances in the confusion case. Each arrow points to an image panel on the right side for displaying images belonging to each confusion case. These image panels are vertically aligned side by side, helping users observe the differences in the pattern of instances in different confusion cases, especially as to how images predicted as negative class (two top panes) vs. positive class (two bottom panes) are different from each other. 

\begin{figure}[!ht]
    \centering
    \includegraphics[width=0.95\columnwidth]{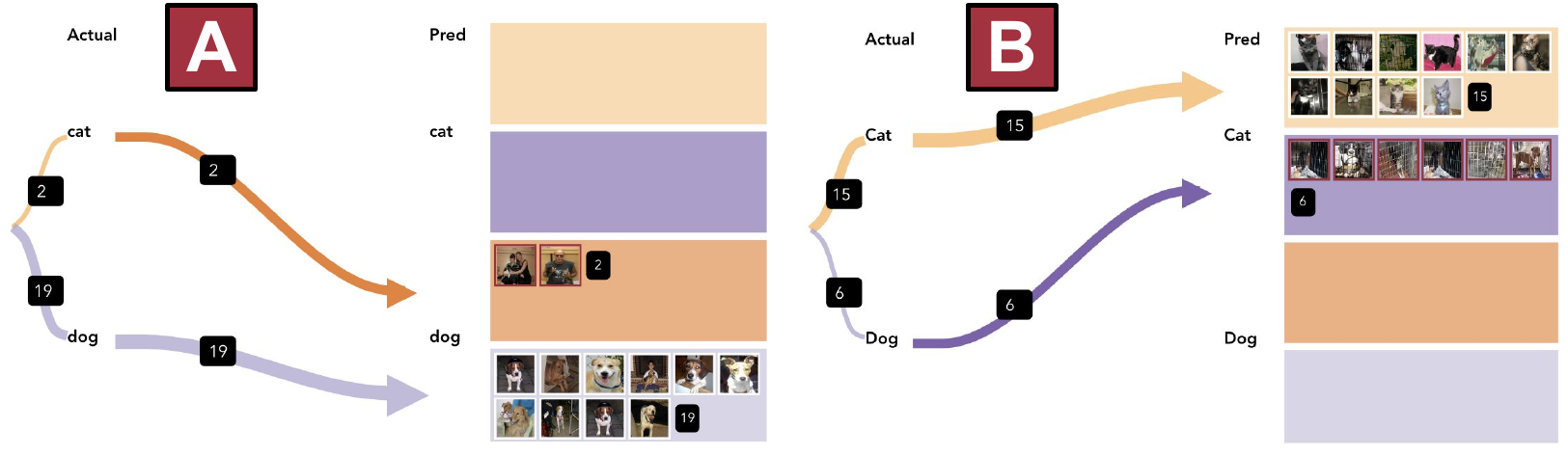}
    \caption{\label{fig:contrastive-analysis-view}
    \textbf{Two examples of \contrastiveview.} \contrastiveview allows users to inspect how a group of images in a similar pattern were learned to be a pattern of either class, (A) two cat images with person misclassified as dogs are identified as similar to a group of dog images with yellow fur or person. (B) A group of cat and dog images with cages are all predicted as cats. 
    }
\end{figure}

\subsection{Segment View}
\label{sec:segment-view}
\segmentview (Fig. \ref{fig:system}D) supports the segmentation of images and allows users to select a set of semantically coherent segments to be defined as a concept. Once identifying specific concepts from images in \contrastiveview, users can select one or two image panes among four and click the segmentation button to generate segments. Once generated, all segment vectors are projected and visualized in a 2D scatter plot (Fig. \ref{fig:system}D-i) as rectangle glyphs with their color indicating based on confusion cases. Once a segment rectangle is hovered over, a tooltip highlights the k nearest segments to allow users to select a group of segments (k=10) that are visually coherent. When a segment is clicked, the nearest neighbors are selected together in the panel (Fig. \ref{fig:system}D-ii). This panel for defining a set of segments as a concept is editable in two ways: users can delete undesirable segment images by clicking or can type in the name of concept.

\subsection{Concept Inspection View}
\label{sec:system-concept-inspection-view}
\conceptinspectionview (Fig. \ref{fig:system}F) provides the visual summary of two aspects of systematic errors, association, and attribution (details in Section \ref{sec:introduction}), for all concepts defined by users in the system to allow users to inspect the concept association against misclassifications (\textbf{T3}). This view consists of two sub-components: \conceptassociationplot and \conceptdetailview. 

First, \conceptassociationplot (Fig. \ref{fig:system}F-i) visualizes two aspects of concept associations, association, and attribution in a 2D scatterplot. In the plot, each concept is represented as a vertically aligned pair of two circles, each representing the degree of false positive (FP) as an upper circle, and false negative (FN) as a lower circle with its size indicating the number of instances in FP and FN. These circle pairs are located in a 2D scatter plot indicating two concept associations in the $x$ and $y$-axis: First, the between-class disparity of concepts (details in Section \ref{sec:method-between-class-disparity}) is represented as the $x$-axis. Second, the concept influences towards two misclassifications (FP and FN) (details in Section \ref{sec:method-concept-influence}) are presented in the $y$-axis, consisting of the top half for the concept influence towards FN and the bottom half for FP. As circles are placed further from the center (towards the top (FP) or bottom (FN)), it indicates higher influence towards misclassifications.

\conceptdetailview (Fig. \ref{fig:system}F-ii) is designed to illustrate the details of two aspects of associations and attributions for a certain concept and to display the most associated instances. In the horizontally aligned panels, two aspects of associations visualized in \conceptassociationplot are illustrated as a form of questions, ``What did the model learn?" and ``What did the model fail?". In each panel, the class-level concept associations (details in Section \ref{sec:method-between-class-disparity}) are presented as bar charts with a list of the most concept-associated instances.

\subsection{Debias View}
\label{sec:debias-view}
\debiasview (Fig. \ref{fig:system}G) enables users to confirm how the associations between concepts and certain classes can be mitigated via the debiasing method (\textbf{T4}) (Section \ref{sec:method-debiasing}). In a line chart layout, each concept is represented as a line showing how biases can be gradually decreasing. The chart shows the ratio of remaining associations compared to one before debiasing presented as 1.0 in the $y$-axis (Equation \ref{equation:remaining-bias-ratio}) as the number of concept-associated instances in the $x$-axis is increasingly debiased. We also provide a slider interface where users can determine their tolerance in the amount of instances to be debiased, to help them gauge the trade-off between the number of instances and the degree of biases. As a user adjusts the slider between 0 and 1 (from less to more number of instances), the optimal point based on user's preference is adjusted and highlighted with its red stroke. In the below of \debiasplot, we summarize how the between-class disparity changed before and after removing biases in the form of a textual description to better illustrate the effect of debiasing spurious associations.

\section{Evaluation}
\label{sec:evaluation}

To validate the effectiveness of our system and methods, we conduct the evaluation in three ways. First, we conduct a quantitative evaluation on our two statistical methods, combined concept association (Section \ref{sec:method-concept-association}) and debiasing spurious concept association (Section \ref{sec:method-debiasing}). Second, we demonstrate the utility of \name with two case studies, expert interview and user study. Third, we conduct a controlled user experiment to evaluate how ML practitioners can successfully perform the workflow of inspecting systematic errors using \name. Throughout the evaluations, we use two different benchmark datasets to demonstrate the application of our approach as follows:

\begin{itemize}
    \item \textbf{Cats \& Dogs dataset (skewed/original)} \cite{cats_dogs}: In the first setting, we train two versions of the image classifier over original or skewed Cats \& Dogs dataset. First, the original dataset with 2,600 cats and dogs images was used as a setting to identify systematic errors of a model trained. On the other hand, the skewed dataset, where spurious associations from the original data were intensified by removing concept-related images in the minor classes in the training set, was also used to 1) simulate systematic errors to have more diverse concepts in the analysis and 2) have a set of ground truth concepts that are intentionally attribute systematic errors to validate how those concepts were effectively discovered and mitigated via our methods and system. These concepts include grass, person, jean/skyblue (biased towards dogs), cage, and red/pink objects (biased towards cats). We use the skewed setting to present a case study to demonstrate the usage scenario (Section \ref{sec:case-study-1}) and (3) to conduct a user study to validate the usefulness of the system with ML practitioners (Section \ref{sec:user-study}).
    \item \textbf{Pascal Voc dataset} \cite{pascalvoc}: This dataset demonstrates a multi-classification scenario with 20 classes including animals and objects to present a case study in Section \ref{sec:case-study-2}. We showcase systematic errors over four classes (bird, train, aeroplane, chair, and car) over five concepts (grass, sky, cloud, red\&blue, and straight).
    \item \textbf{CelebAMask-HQ} \cite{celeba-hq}: This dataset includes face images with 40 annotated attributes. We train a classifier on two gender classes (female and male) to demonstrate systematic errors over five classes (smiling, eyeglasses, wearing hat, not smiling, and blond hair).
\end{itemize}

\textbf{Implementation.} Throughout the settings, we implement the image classifications using Resnet50, a convolutional neural network with residual components consisting of 16 blocks of bottleneck layers in a total of 50 individual layers. We employ a pre-trained ResNet50 model \cite{he2016deep} over ImageNet dataset from Pytorch\footnote{https://pytorch.org/} as a base model and train it with either Cats \& Dogs or Pascal Voc dataset described above. Throughout the analysis, we choose 14th bottleneck layer located in the later part of the layer structure with the 8x8x2048 dimension of internal representations of instances and segments.
\section{Quantitative evaluations}
\label{sec:quantitative-evaluations}

We conduct two quantitative evaluations to validate two methods: (1) Combined concept associations (Section \ref{sec:method-concept-association}) and (2) Debiasing spurious associations (Section \ref{sec:method-debiasing}). First, we evaluate how combined concept association precisely identifies concept-associated instances. We compare our method with raw concept association as a baseline by measuring the precision (P@5, 10) over misclassifications in four settings. The result summarized in Table \ref{table:experiment-result-1} shows that our method overall outperformed the baseline method in finding the concept association. 

\begin{figure}[!ht]
    \centering
    \includegraphics[width=\columnwidth]{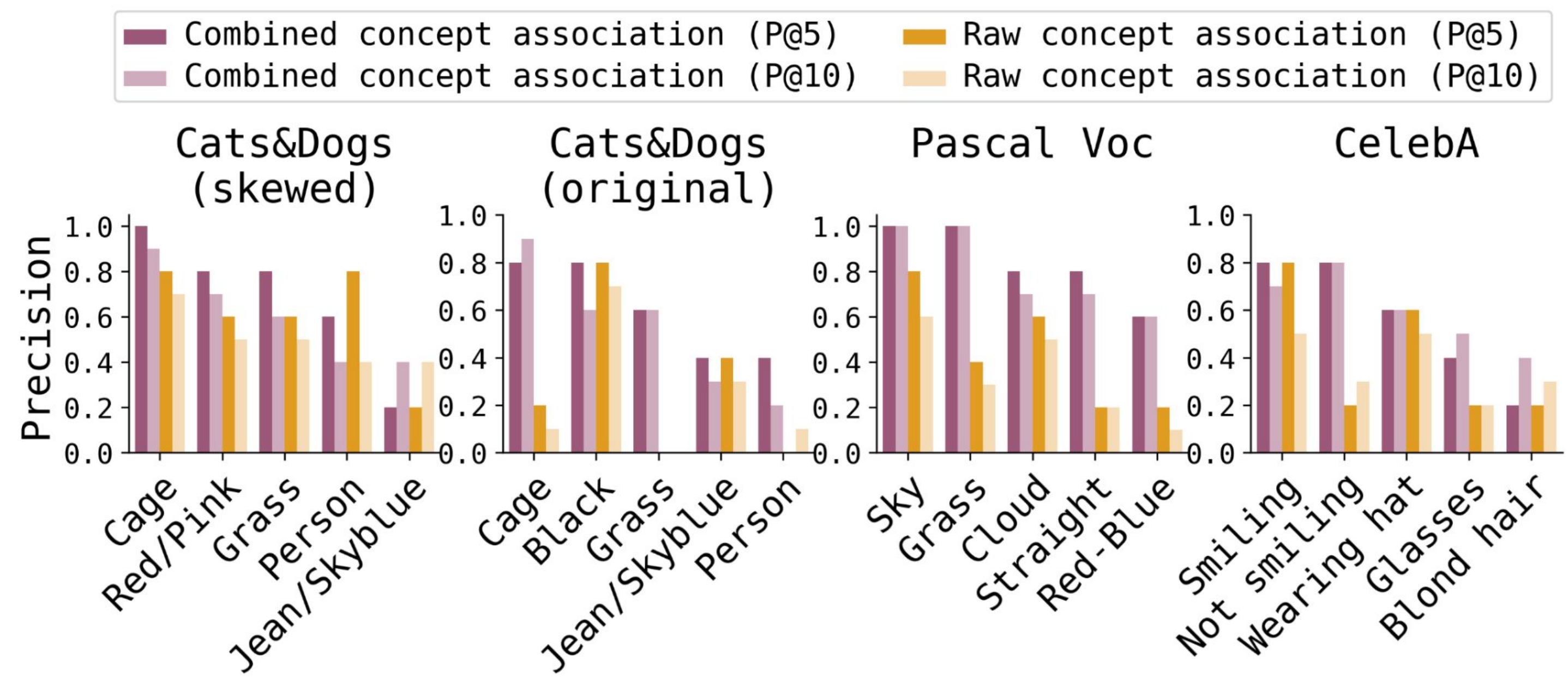}
    \vspace{-13.5pt}
    \label{table:experiment-result-1}
    \caption{\label{table:experiment-result-1} The experiment results in comparing combined concept associations (ours) and raw concept association (baseline). It shows that our method performs better in retrieving concept-associated instances.}
    \vspace{-0.25em}
\end{figure}

Second, we evaluate how the debias method (details in Section \ref{sec:method-debiasing}) can help mitigate predictive errors. In the experiment, we selected top-associated concepts and classes for each setting that were the most selected by participants in the user study (for the skewed version of Cats\&Dogs) or by authors based on \debiasview (for the rest of settings). The effectiveness of mitigation is evaluated against randomly selected instances as a baseline. Table \ref{table:experiment-result-2} summarizes the statistics about the associations, debias statistics, and performance before and after debias. It shows that debiasing spurious associations within two concepts improves the performance, compared to a random instance set which in contrast lead to the degradation in accuracy.

\begin{table}[]
\begin{tabular}{l|l|l|l|l}
\hline
                                                                                 & \textbf{\begin{tabular}[c]{@{}l@{}}Association\\ (Concept\\ /Class)\end{tabular}} & \textbf{\begin{tabular}[c]{@{}l@{}}\# Instances\\ to debias\\ (\% Bias\\ mitigated)\end{tabular}} & \textbf{\begin{tabular}[c]{@{}l@{}}Before\\ debias\\ (Acc)\end{tabular}} & \textbf{\begin{tabular}[c]{@{}l@{}}After\\ debias\\ (Acc)\end{tabular}} \\ \hline
\multirow{3}{*}{\begin{tabular}[c]{@{}l@{}}Cats\&Dogs\\ (Skewed)\end{tabular}}   & Grass/Dog                                                                         & 200 (50.7\%)                                                                                      & \multirow{3}{*}{0.918}                                                   & 0.932                                                                   \\
                                                                                 & Person/Dog                                                                        & 100 (13\%)                                                                                        &                                                                          & 0.922                                                                   \\
                                                                                 & Random                                                                            & 200 (-)                                                                                           &                                                                          & 0.915                                                                   \\ \hline
\multirow{3}{*}{\begin{tabular}[c]{@{}l@{}}Cats\&Dogs\\ (Original)\end{tabular}} & Grass/Dog                                                                         & 400 (10.9\%)                                                                                      & \multirow{3}{*}{0.880}                                                                   & 0.887                                                                   \\
                                                                                 & Cage/Cat                                                                          & 200 (12.9\%)                                                                                      &                                                                          & 0.887                                                                   \\
                                                                                 & Random                                                                            & 200 (-)                                                                                           &                                                                          & 0.878                                                                   \\ \hline
\multirow{3}{*}{CelebA}                                                          & Smiling/Female                                                                    & 300 (19.7\%)                                     & \multirow{3}{*}{0.906}                                                                    & 0.910                                                                   \\
                                                                                 & Eyeglasses/Male                                                                   & 600 (41.7\%)                                                                                      &                                                                          & 0.911                                                                   \\
                                                                                 & Random                                                                            & 300 (-)                                                                                           &                                                                          & 0.900                                                                   \\ \hline
\multirow{2}{*}{Pascal Voc}                                                      & \begin{tabular}[c]{@{}l@{}}Straight\\ /Aeroplane*\end{tabular}                    & 50 (49.6\%)                                                                                       & \multirow{3}{*}{0.813}                                                                & 0.820                                                                   \\
                                                                                 & Random                                                                            & 50 (-)                                                                                            &                                                                          & 0.811                                                                   \\ \hline
\end{tabular}
\label{table:experiment-result-2}
\caption{\label{table:experiment-result-2} The experiment results in evaluating the effect of debias method in improving predictive performance. The reduced associations between concept-class from concept-associated instances were evaluated to improve the performance (*From binary case between Train/Aeroplane class).}
\vspace{-5pt}
\end{table}
\section{Usage scenarios}
\label{sec:case-studies}

\subsection{Usage scenario I: Identifying concept associations from a biased dataset}
\label{sec:case-study-1}
In the first case study, we present a use case scenario to demonstrate how biased concepts can be detected with the use of \name under the skewed Cats \& Dogs setting described in Section \ref{sec:evaluation}. We aim to validate the effectiveness of the system with diverse ground truth concepts that were intentionally designed to induce systematic errors. We interviewed two experts E1 and E2, who have expertise in predictive modeling for the research in statistics and chemical engineering. 

\textbf{Diagnose the potential of systematic failures (T1)}. When they started navigating the system, E1 noticed that \performancechart gave a summary of two types of misclassifications, ``\textit{It gives me a perspective of how I can further differentiate two types of misclassifications and figure out how the degree of unknown-unknowns are severe.}'' After confirming a high portion of unknown-unknowns, they further investigated how predictions are distributed over four confusion cases in \confusionmatrix (Fig. \ref{fig:system}A-ii). E2 immediately captured that the matrix showed higher degree of misclassifications in two classes with circles with darker and lighter red, ``\textit{I can easily notice I need to take the case of misclassified cats into account more seriously}''. It made him decide to closely examine the possibility of systematic errors in cat class.

\begin{figure}[!ht]
    \centering
    \includegraphics[width=0.95\columnwidth]{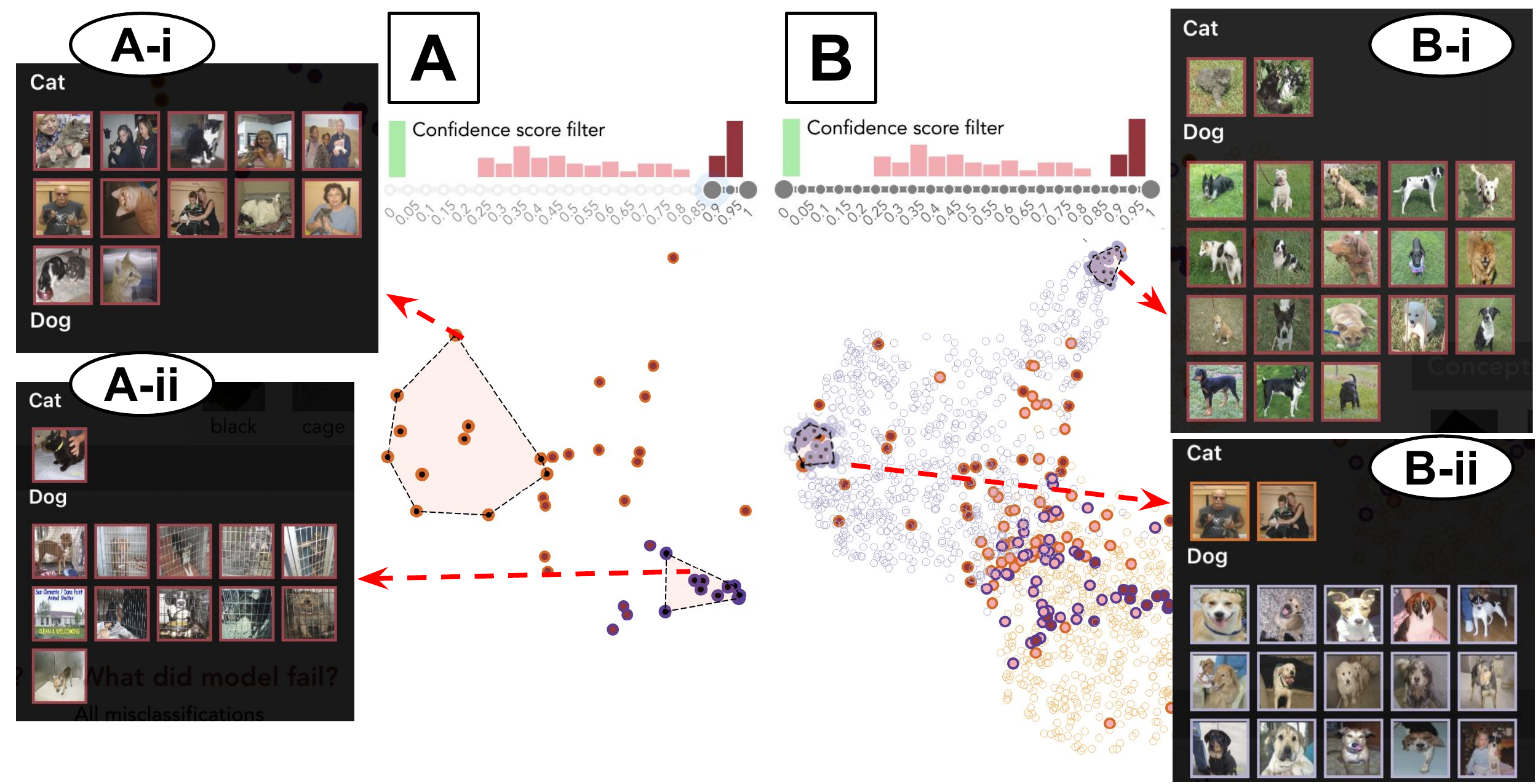}
    \vspace{-7.5pt}
    \caption{\label{fig:case-study-1}
    \textbf{Identifying and contrasting cases for hypothesizing concept associations.} \confidencescorefilter is adjusted to allow users to examine (A) unknown-unknowns exclusively or (B) all true and false instances. 
    }
    \vspace{-7.5pt}
\end{figure}

\textbf{Filter, contrast, and hypothesize} (\textbf{T2}). To find why cats were falsely predicted with high confidence than dogs, E1 used \instancespace and adjusted confidence score filter to exclusively examine misclassified cats with high confidence (Fig. \ref{fig:case-study-1}A). When hovering over two regions of misclassified cats of his interest, he identified that cat images with \textbf{person} and \textbf{grass} background appeared in the tooltip (Fig. \ref{fig:case-study-1}A-i, B-i). When extending the filter to investigate both true and false predictions (Fig. \ref{fig:case-study-1}B), he found that those misclassified cats were mixed up with a number of true dog images. Several cat images with person were in the middle of dog images with yellow or white furs (Fig. \ref{fig:case-study-1}B-ii). This led him to hypothesize that persons’ skin were misunderstood as furs with bright colors by the classifier. E1 also identified there is a subregion where a number of grass images were grouped together. A click showed that those images appearing in \contrastiveview were all found to be predicted as dogs, including a few cat images falsely predicted as dogs. On the other hand, E2 paid attention to a region where images were a group of misclassified dog with \textbf{cage} (Fig. \ref{fig:case-study-1}A-ii). It made him hypothesize the spurious association between \textbf{cage} concept and cat class.

\textbf{Validate concept associations (T3)}. 
To validate the concept association, they began to explore \conceptassociationplot (Fig. \ref{fig:system}F-i). In the plot, a total of seven concepts was shown with four automatically detected concepts and three additional concepts created by E1 and E2 separately, and four other predefined concepts that were also added by the researcher to let them explore a variety of concept influences. They captured at first sight that \conceptassociationplot showed a clear pattern between two types of associations (Fig. \ref{fig:system}F-i). E2 mentioned that, ``\textit{I can see that there is a positive association along two axes. For example, like a grass concept, the concepts biased towards dogs tend to be influential towards misclassified cats. And other concepts biased towards cats like black concept at the leftmost part have the opposite pattern}''. E2 first looked at \textbf{cage} concept to verify his first hypothesis. When selecting \textbf{cage} concept in \conceptassociationplot, it was attributed similarly to both misclassified cats and dogs. But in \conceptdetailview, the top-associated images with cage contained several types of objects such as \textbf{tiles}, \textbf{leashes}, \textbf{walls}, or \textbf{floor rugs}. Those objects were in their stripe or grid-like patterns in common. In the dog images, leashes noticeablely appeared. It indicated that, despite cage concept was intentionally biased to be a pattern of cats in our setting, the model learned various similar objects as a consistent pattern.

E1 previously made a hypothesis on \textbf{person} concept, so he clicked on it in \conceptassociationplot, whose detailed concept associations appeared in \conceptdetailview (Fig. \ref{fig:system}F-ii). In the left panel of \conceptdetailview, he found in the top-associated instance list that his observation on \instancespace was consistent with patterns in images containing cats with yellow furs, which meant the most associated instances towards person concept were white and yellow cats, then it led to the misclassified cats with person or hands. On the other hand, for grass concept, \conceptdetailview (Fig. \ref{fig:system}F-ii) showed a number of grass-associated images with a bar chart indicating a higher degree of grass concept associated with dogs than with cats. 

\textbf{Mitigate spurious associations via debiasing (T4).} After the investigation, E1 decided that grass concept is the most noticeable in spurious concept associations, ``\textit{I think \textbf{grass} concept is a background concept and highly biased toward dogs, so I want to debias it.}'' When E1 clicked debiasing button, the debiasing effect of grass concept was highlighted in \debiasview with 413 concept-associated instances  at the maximum as candidates of debiasing process. The plot also highlighted the optimal number of instances based on the preference over tolerance slider. E1 said, ``\textit{I think debiasing 200 instances are better because where the decreasing trend starts to flatten}''.

\begin{figure*}[!ht]
    \centering
    \includegraphics[width=0.95\linewidth]{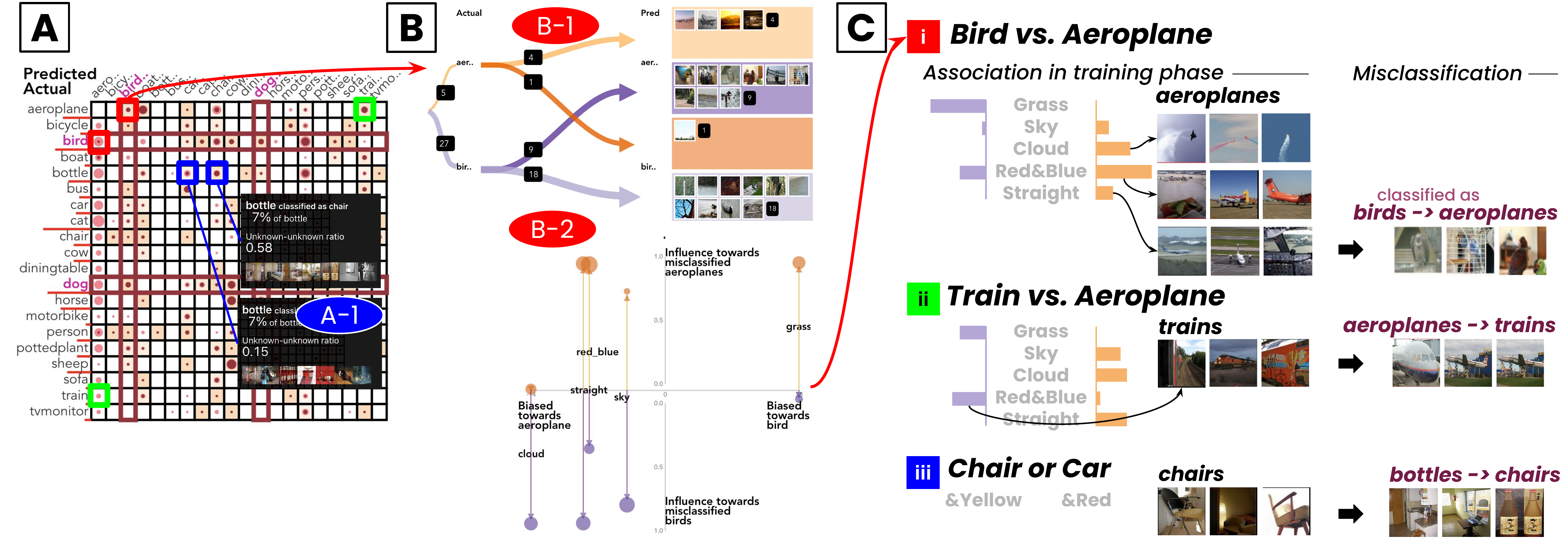}
    \caption{\label{fig:case-study-2}
    \textbf{Exploring the spurious associations in Pascal Voc dataset using \name.} In the second case study, (A) Jane found in \confusionmatrix that some classes had more unknown-unknowns (indicated as red bars on the row labels) and began to take a closer look at three competing pairs of classes. (B) Using \contrastiveview (top) and \conceptinspectionview (bottom), she hypothesized five concepts (shown in C-i). (C) Three competing pairs of classes (i-iii) result in spurious associations between  concepts and certain class, leading to misclassifications in the opposite class.}
\end{figure*}

\subsection{Usage scenario II: Exploring spurious associations from real-world multiclassification images}
\label{sec:case-study-2}
In the second scenario, we present a case study of  exploring spurious associations in the second real-world dataset using \name (Fig. \ref{fig:case-study-2}). In the scenario, we take an example of a data scientist named Jane, who wants to inspect systematic errors in a multi-class image classification tasks with Pascal Voc dataset described in Section \ref{sec:evaluation}.

\textbf{Narrow down to regions of systematic errors (T1)}. After the system was initially loaded, she checked the higher proportion of unknown-unknowns in \performancechart and found in \confusionmatrix (Fig. \ref{fig:case-study-2}) that some classes such as aeroplane, bird, cat, and dog classes had more number of unknown-unknowns indicated as the length of red bars. By hovering over specific confusion cases in \confusionmatrix, she observed that misclassified images in some confusion cases commonly include some semantic concepts. For example, two groups of bottle images misclassified as either car or chair (Fig. \ref{fig:case-study-2}A-i) showed a stark difference in patterns (details below). After exploring coherent patterns to detect the potential regions of systematic errors, she decided to closely inspect three pairs of classes (bird/aeroplane, train/aeroplane, and chair/car/bottle, as highlighted with red, green, and blue in Fig. \ref{fig:case-study-2}A).

\textbf{Explore and validate concepts and the spurious associations over pairs of classes (T2, T3)}. First, she paid attention to the binary case of bird and aeroplane by clicking two class names in \confusionmatrix. When she moved on to \instancespace and selected a region containing a group of bird images falsely predicted as aeroplanes and other images as well, \contrastiveview diverged (Fig. \ref{fig:case-study-2}B-1) them into two different groups of bird images. In two different panes (colored light and dark purple), a group of the correctly classified bird images were in the forest or tree background, whereas in a group of falsely predicted bird images, birds appeared in a fence, cage, or shelter that share \textbf{straight} or \textbf{grid}-like patterns. After she further explored other regions of aeroplane \& bird, she defined \textbf{grass}, \textbf{sky}, \textbf{cloud}, \textbf{red \& blue stripe}, and \textbf{straight} in \conceptlist. In \conceptassociationplot (Fig. \ref{fig:case-study-2}B-2), four concepts except for grass were located in the left side of the plot, indicating they were biased towards in aeroplanes. As shown in Fig. \ref{fig:case-study-2}C-i, the top-associated images included aeroplanes leaving contrails over the sky (as cloud), in the middle of the runway airport (as straight), or with red \& blue prints (red\&blue stripes). She was wondering how the aspect of associations differs in the case of aeroplanes \& trains (Fig. \ref{fig:case-study-2}C). She clicked two classes in \confusionmatrix, explored the associations with the same concept set. Interestingly, the \textbf{red\&blue stripe} concepts were biased towards train class in this case. As summarized in Fig. \ref{fig:case-study-2}C-ii, the top-associated images were mostly trains with vivid prints on their bodies that are zoomed in thus take the most spaces in the images.

While further exploring \confusionmatrix (Fig. \ref{fig:case-study-2}A-i), she found that two sets of bottle images misclassified as cars or chairs were mostly colored as \textbf{red}, or \textbf{yellow} or placed in a house with many furnitures respectively, meaning the associated patterns were bound to the opposite class. She hypothesized multiple colors including \textbf{yellow}, \textbf{red}, \textbf{blue}, and \textbf{gray}, and \textbf{wood} as brown. she could find in the system that chair class was highly biased towards \textbf{yellow} and \textbf{wood}. As shown in Fig. \ref{fig:case-study-2}C-iii, the top-associated images in \textbf{yellow} concept in the training set were mostly chairs made of woods or placed in the wood floor exhibiting yellow colors.

\section{User Study}
\label{sec:user-study}

To evaluate the effectiveness and utility of \name, we conduct a controlled user study to compare ESCAPE with a baseline system. In the study, we set up the baseline system with a setting of visualizations and interactions consisting of two views: a confusion matrix visualization and a list of images with three information including its true and predicted class, and model confidence. While this is a workaround due to the inability to run interactive tools from relevant studies (i.e., tools are not available or requires auxiliary data), the baseline system represents a simple and standard setting that ML practitioners commonly use when debugging misclassifications in practice. The design of our controlled experiment also can be viewed as comparing two versions of our system, between the system version A only with \confusionmatrix and the version B with all functionality of \name, which allows us to test the utility of other key components in \name such as \instancespace, \contrastiveview, \conceptassociationplot, and \debiasview.  

\textbf{Study overview.} In the study, we recruited 22 participants (15 male and 7 female) via university email listserv. All participants had machine learning or statistical analysis experiences who are studying Information and Computer science or Statistics (Bachelor’s: 3, Master’s: 6, Doctorate: 13). Each participant was assigned to use either ESCAPE or baseline system for us to conduct a between-subject experiment. In an hour-long session, participants were given 15 minutes of introduction to the study and the walk-through of the system, and were tasked with thinking aloud to perform tasks and answer questions summarized in the below section. Finally, participants were asked to complete usability questions using a 5-point Likert scale. Participants conducted the study via either an online session using video conferencing or in-person session and were rewarded with \$15 upon the completion of the study.

\textbf{Questions and Tasks.} In the study, participants were asked to perform four tasks designed to test the utility of workflow (T1-T4) by answering Q1-Q8 in Cats\&Dogs image classification setting illustrated in Section \ref{sec:evaluation}: \\

\textbf{T1.} Diagnose: Inspect the degree of misclassifications (Q1-Q3) \\
\indent \textbf{T2.} Identify: Discover and hypothesize the associations between concepts and target classes (Q4) \\
\indent \textbf{T3.} Validate: Rank the given concepts in the order two aspects of spurious associations (Q5-Q6) \\
\indent \textbf{T4.} Mitigate: Determine a concept to debias and the number of concept-associated instances to mitigate the spurious association (Q7-Q8) \\

For T1 and T2 (Q1-Q4), we provided a system without any predefined concepts to let participants find out and hypothesize concepts based on their own observations. In the second half of the study (for T3 and T4, Q5-Q8), participants were given seven predefined concepts and are evaluated in terms of how the given system helped them capture the concept associations that caused misclassified cats.

\begin{table}[]
\begin{tabular}{@{}lll@{}}
\toprule
\textbf{Concept - Class} & \textbf{ESCAPE} & \textbf{Baseline} \\ \midrule
Cage - Cat               & 8/11 (72.7\%)   & 2/11 (18.2\%)     \\
Grass - Dog              & 7/11 (63.6\%)   & 1/11 (9\%)        \\
Person - Dog             & 6/11 (54.5\%)   & 4/11 (36.3\%)     \\
Red/Pink object - Cat    & 1/11 (9\%)      & 0/11 (0\%)        \\
Jean/Skyblue - Dog       & 0/11 (0\%)      & 1/11 (9\%)        \\ \bottomrule
\end{tabular}
\label{table:user-study-result-1}
\caption{\label{table:user-study-result-1} The count of identified associations between concept-class pairs by \name and baseline groups in the user study (T2).}
\vspace{-5pt}
\end{table}

\textbf{Results.} The result in Table \ref{table:user-study-result-1} shows that participants in ESCAPE successfully performed T1-T3 better than baseline users. In Diagnosis task (T1), \name users accurately diagnosed the degree of misclassifications at instance (11/11, 100\%) and class level (11/11, 100\%) compared to baseline users (5/11 (45\%), 1/11 (9\%) for each). In the task of identifying the associations (T2), more portion of \name users accurately detected the association of three concepts (cage, person, and grass) associations with a class (Table \ref{table:user-study-result-1}). On the other hand, two other concepts were identified by only a few participants from both groups. This may be due to less noticeable patterns of two concepts appearing in the test set as these concepts were the least biased towards certain class (details in Table \ref{table:experiment-result-1}). In addition to these five concepts, both groups explored a variety of concepts in their task. These concepts included 26 unique concepts across diverse types such as background (inside, shelter), appearance (triangle ear, tongue), color, size and pose (small, facing front) or combination of features (black furs without person). Lastly, according to the result of subjective ratings in a 5-point Likert scale to test the usefulness, interactiveness, and usability of two systems and the one-tailed two sample t-test, participants found \name was more useful (\name: $\mu$ = 4.91, $\sigma$ = 0.3, baseline: $\mu$ = 3.8, $\sigma$ = 0.94, $p$=0.001) and interactive (\name: $\mu$ = 4.33, $\sigma$ = 0.82, baseline: $\mu$ = 3.7, $\sigma$ = 1.21, $p$=0.039) than baseline system. However, \name was evaluated as not easier to use than baseline due to its higher complexity with various components although the difference was not statistically significant (\name: $\mu$ = 4.08, $\sigma$ = 0.89, baseline: $\mu$ = 4.4, $\sigma$ = 0.52, $p$=0.081).

\textbf{Behavioral patterns.} In addition to study results, we analyze the behavioral patterns of participants in inspecting systematic errors while using two systems. Three findings highlight how the design of \name promotes a careful inspection of systematic errors and mitigation of biases. \\

\textbf{Finding 1. \name users exhibited an increased awareness of the concept of model confidence.} During the study, four users reported their preferences to investigate unknown-unknowns, ``\textit{I want to look at high confidence error first,}'' ``\textit{I wasn’t aware of how errors can be categorized by model confidence but I can see how it is critical.}'' These four users commonly investigated instances in the order of unknown-unknowns, known-unknowns, and true predictions by adjusting the confidence score filter. They also mentioned their increasing awareness of unknown-unknowns and the importance of inspecting model confidence during the study.

\textbf{Finding 2. Practitioners tend to contrast predictive cases to find the patterns of misclassifications.} In T2, we could observe that 8 of 11 baseline users actively compared at least one or two pairs of confusion cases. Most of them noticeably paid attention to comparing FP with counterfactual cases whose pairs are both true cats (TN vs. FP: 6/11 (54.5\%) users) or ones predicted as dogs (TP vs. FP: 5/11 (45.5\%) users) that are minimally different. While this proves how ML practitioners prefer to compare different cases to find out patterns, it also gave the rationale behind the design of our \contrastiveview that incorporated the way practitioners by nature analyze the pattern.

\textbf{Finding 3. \name users tend to have different decision criteria for determining how to best debias.} In T4, we noticed that participants determined the number of instances to debias with different decision criteria. While most of them (8/11, 45\%) chose the optimal point (i.e., elbow point), others preferred to choose the threshold that maximally removes the biases (2/11, 18\%) or to debias the maximum number of instances (1/11, 9\%). This showed the debias tolerance slider in \debiasplot to be effective in incorporating their preference and recommending the optimal point.

\vspace{-10pt}
\section{Discussion}
\label{sec:discussion}

\subsection{System utility} 
\textbf{Usability of components.} The feedback from study participants and experts was generally positive. Most of them found that the use of \name informed them of the otherwise hidden associations between concepts and classes. In the user study, \contrastiveview, and \instancespace were most often used by participants to iteratively make hypotheses. Among multiple components, \conceptassociationplot was the most appreciated by 5 participants. They mentioned that it helped them to make sense of the two interrelated aspects of associations that practitioners are typically unable to quantify. Also, two participants mentioned about the informativeness of \debiasview , ``\textit{it helps to understand how much data cleaning and training process should be done right, which is very critical to know.}’’ In the expert interview, two practitioners found \name to be useful in performing the proposed workflow by successfully identifying the intentionally biased concept-class pairs. The qualitative evaluations showed our statistical methods were effective in identifying and mitigating spurious associations compared to baseline settings.

\textbf{Concept-level systematic error debugging tool.} There are other tools available in the area of concept-based interpretability and systematic error detection. Here, we summarize when and how \name is better capable of supporting the following tasks:
\begin{itemize}
    \item Debugging-focused workflow and visualization: Compared to other visual analytic tools for concept-level interpretability such as ConceptExtract \cite{zhao2021humanintheloopextraction} or ConceptExplainer \cite{ConceptExplainerUnderstandingMentalModelDeep}, ESCAPE visually highlights and validates spurious associations between concepts and misclassifications.
    \item Human-in-the-loop workflow: While many automatic methods support the discovery of sub-regions as a part of required tasks, we highlight the comprehensive workflow to showcase how practitioners can leverage their capability of identifying spurious associations in systematic errors to generate and testing their hypotheses and precisely debug systematic errors.
    \item Quantifying the global influence of concepts: Our system is capable of quantifying the association and attribution of concepts to their impact on misclassification at the \textit{global} level.
    \item No need for auxiliary data: Compared to some studies leveraging additional data to capture the associated concepts such as text embedding \cite{DominoExtractingComparingManipulating} or user-generated report \cite{DiscoveringValidatingAIErrorsCrowdsourced}, our tool does not require auxiliary data or segments necessary to define concepts. 
\end{itemize}

\subsection{Human behavior in concept discovery}
We summarize our findings in the user study on human behavior and capability in discovering concepts. 

\textbf{Concept diversity and heuristic thinking.} During the user study, we found that participants are able to discover diverse concepts using heuristic thinking and common sense. Throughout the sessions, 26 unique concepts encompassing 11 different types such as object, pattern, size and pose, background, and location were discovered. We observed that participants often found concepts with thinking aloud in leveraging common sense (e.g., ``Dogs are usually inside than outside, so dogs may be more associated with grass.'' or ``Cats are inside and likely to be in a cage or with stuff at home.''). Such heuristic-driven analysis can be leveraged to filter out false concepts that are too deviated from common sense.

\textbf{Trade-off between precision and recall.} From the user study, we found that humans are better at precisely capturing concepts (precision: 0.75) despite achieving lower recall (recall: 0.39). For some concepts that rarely appear (e.g., red/pink or jean/skyblue in Table \ref{table:user-study-result-1}), participants were not able to hardly identify them. While automatic discovery is better at collecting significant pieces of underperformed data, but these methods still fail to recover over ~60\% of coherent slices \cite{DominoExtractingComparingManipulating} and also lack precisely capturing concepts \cite{DominoExtractingComparingManipulating, SpotlightGeneralMethodDiscoveringSystematic}. This sheds light on collaboration in automatic and human-driven methods to capture spurious associations better. For example, it is desirable to introduce some automatic methods in Identification stage in our system and accommodate human supervision in guiding subspace discovery. \\

\subsection{Generalizability}
In this section, we discuss the generalizability of our workflow and system toward data types, tasks, and applications.

\textbf{Data type}: The system interface was built to support the comprehension and mining of concepts in image data with image segmentation and contrasting images, but the framework can serve a wider range of data representations such as texts where fine-grained units as concepts are human-interpretable (vs. time-series or graphs whose subpatterns do not have semantic meanings).

\begin{figure}[!ht]
    \centering
    \includegraphics[width=.95\columnwidth]{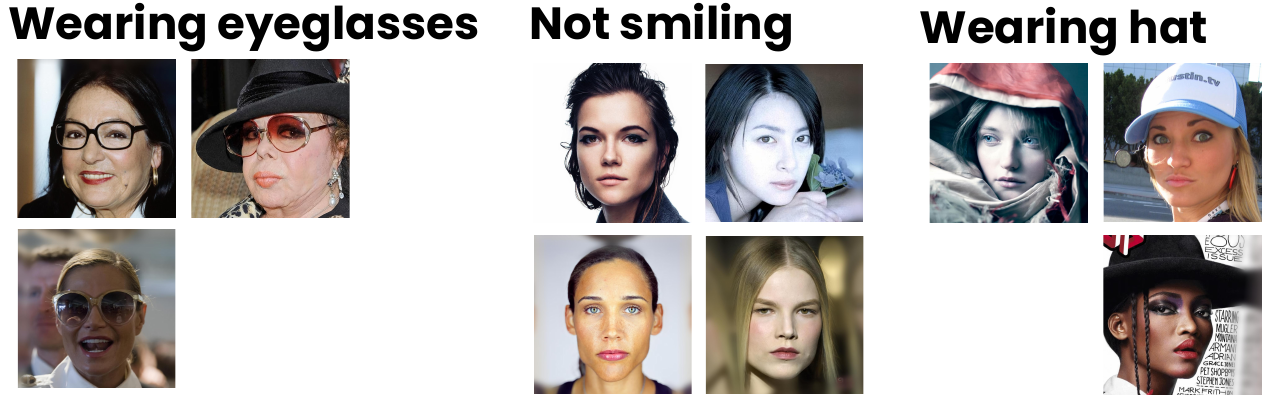}
    \caption{\label{fig:case-study-3}
    \textbf{Examples of females misclassified as males with certain attributes in gender classification from CelebA-HQ dataset.}}
    \vspace{-5pt}
\end{figure}

\textbf{Applicability to high-stakes scenario.} As illustrated in Section \ref{sec:introduction}, there are increasing number of AI applications impacting individuals to a greater extent. Our system can be also applicable to detecting spurious associations in the applications such as medical imaging (i.e., are there any missing cases of symptoms that can lead to diagnosis?). This was partially demonstrated with CelebA dataset whose classification can impact subgroups that have certain attributes as illustrated in Fig. \ref{fig:case-study-3} such as female with eyeglasses, hats, or no smiling.

\textbf{System and workflow}: While our system and workflow were demonstrated in discovering systematic errors and mitigating biases, these can generalize to other tasks and methods. For example, \name can be leveraged to expore concept interpretability especially for subspaces of interest in general. In addition, relative concept association can be leveraged to identifying concept-associated instances from in-the-wild cases for extending the dataset. In the mitigation stage of the workflow, active learning approach can mitigate the bias as widely demonstrated in existing work.

\subsection{Limitations and future work} In this section, we summarize the limitation and the possible future work to deliver insights and takeaways in five aspects.

\textbf{Scalability.}
Several components in \name (e.g., \instancespace and \segmentview) allows users to identify and contrast instance-level patterns in a test set and define coherent concepts with a group of segments. Nevertheless, the current design has a limited scalability for a larger number of instances and segments. According to our experiments, the system may experience degradation while rendering more than 15,000 visual elements. To get around this issue, the system may extend its ability to randomly filter out some correctly predicted instances to reduce the total number of instances while still loading all misclassification cases.
    
\textbf{Complex interaction between concepts.} During the user study, we also found that some users hypothesized combinations of concepts as a trigger of systematic errors (e.g., blackness but no person). In this case, a model may learn a composite of concepts as patterns of some classes rather than individual concepts separately. While we provided a perspective that multiple concepts in individual instances may influence predictions in a competitive manner, the complex interactions between concepts can be further investigated to see if the influence of concepts over systematic errors are amplified or have canceling-out effects when concepts are interacting with each other.

\textbf{Slice(subclass)-based vs. class(confusion-matrix)-based summary.} Among existing work in discovering systematic errors different in units of summary, namely slice-based (others) and class-based summary (ours), we find several studies with slice-based approach, which aim to summarize systematic errors from the entire instance space rather than per-class subspaces, are beneficial in identifying groups of instances that are semantically coherent but distributed across several classes. On the other hand, class-based summary gives a clear overview of which confusion cases are vulnerable with spurious associations between classes, which are better aligned with classification goal (e.g., male and female). These approaches, therefore, need to be carefully examined depending on the goal of given tasks.

\textbf{Extendability to visualize multi-classes.} The workflow of \name is designed to let users focus on the analysis of a binary case that is particularly vulnerable to systematic errors for two purposes (details in Section \ref{sec:misclassification-diagnosis-view}). The system can be extended to visualize multiple classes at once. An automatic support to detect vulnerable cases will be of help to lesson users’ cognitive support without overwhelming components presented in the system.

\textbf{Better inform the need for mitigation phase.} While our system was mostly appreciated by participants throughout the user study, three of them found \debiasview to be less useful than other components, or were not sure about whether mitigation was necessary. For the future work, the system can enhance the debias module to intuitively guide what it serves, and add visual illustrations to better educate users the consequences of biases in various scenarios when they are not mitigated.

\section{Conclusion}
In this work, we proposed \name, an interactive visualization tool for inspecting systematic errors and spurious concept associations. Our work aimed to promote a pragmatic workflow that help better inform users of the underlying mechanism behind systematic errors, namely spurious associations between undesirable patterns and target classes. To support the workflow beyond the challenges in existing tools, visual components in \name highlight the regions of systematic errors and affords human users to keep track of what undesirable associations were learned in the training process. A suite of comprehensive evaluations showed the effectiveness and utility of our system and methods. Especially, the user study with ML practitioners suggests that \name provides suitable aid for users to identify and hypothesize concepts and their associations with target classes, which was also demonstrated in the analysis of their behavioral patterns.

\begin{acks}
The authors would like to acknowledge the support from the AFOSR awards and DARPA Habitus program. Any opinions, findings, and conclusions or recommendations expressed in this material do not necessarily reflect the views of the funding sources.
\end{acks}

\bibliographystyle{ACM-Reference-Format}
\bibliography{references, references-zotero}

\appendix

\end{document}